\documentclass[research]{lineagepreprint}
\usepackage{layout}
\usepackage{subfig}
\usepackage{hyperref}
\usepackage{multirow}
\usepackage{booktabs} 
\usepackage[normalem]{ulem}

\title{Stage-wise Lineage-aware Dropout for Sparse-View 3D Gaussian Splatting}

\author[1,2,4]{Yan Xing}
\author[1,+]{Zhi Chen}
\author[3,2]{Fanglue Zhang}
\author[2]{Junhong Zhao}

\author[2]{Bing Xue}

\author[2]{Mengjie Zhang}
\address[1]{School of Mathematics, Hefei University of Technology, Hefei 230009, China}

\address[2]{School of Engineering and Computer Science, Victoria University of Wellington, Wellington 6140, New Zealand}

\address[3]{School of Computer Science and Engineering, University of New South Wales, Sydney 2052, Australia}

\address[4]{Joint International Research Laboratory of Spatio-temporal Information and Intelligent Location Services, Guilin 541004, China}

\corremail{2025111357@mail.hfut.edu.cn}

\begin{abstract}
Sparse-view 3D Gaussian Splatting (3DGS) is prone to overfitting because only a limited number of images are available to constrain scene geometry. Dropout is one regularization strategy for alleviating this problem. We introduce the concept of a lineage family to describe a set of Gaussian primitives derived from the same founder through cloning or splitting operations. The founder Gaussians refer to all Gaussian primitives generated during initialization. Descendants within the same family often jointly contribute to the refinement of adjacent local regions. However, dropping individual Gaussians can be less effective due to the compensation effect from neighboring Gaussians for the dropped ones. To address this issue, we propose Stage-wise Lineage-aware Dropout (SLD), which applies lineage-family dropout during the structural regularization stage and switches to individual dropout during the later refinement stage. We further propose Lineage-Balanced Densification (LBD), which operates during densification by constructing an expanded candidate pool and reallocating the densification budget according to the mean gradient scores and existing descendant sizes of different families. Experiments on the LLFF, MipNeRF360, and Blender datasets show that the proposed framework achieves competitive performance in sparse-view reconstruction. 
\end{abstract}

\hypersetup{hidelinks}
\begin{document}

\begin{figure*}[t]
    \centering
    \includegraphics[width=\textwidth]{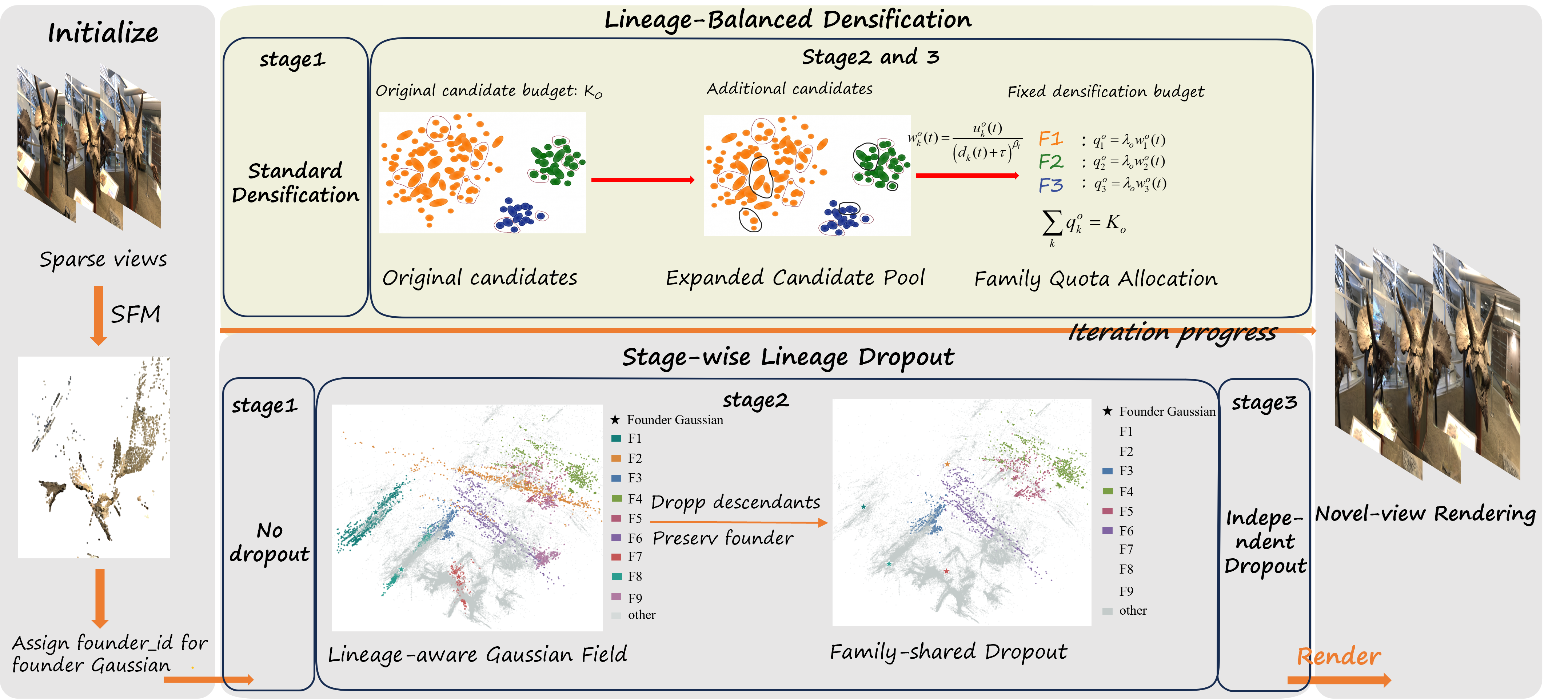}
    \caption{Overview of the proposed lineage-aware Gaussian optimization method. The input views are first processed through SFM initialization, and each initial Gaussian is assigned a founder identity to form lineage families. The SLD component deactivates descendants from selected families. A currently existing generation-0 Gaussian is excluded from this dropout operation. During densification, LBD uses a lower gradient threshold to build an expanded candidate pool and reallocates the densification budget among different lineage families. SLD controls which Gaussians participate in optimization, while LBD selects densification candidates. Both contribute to the final Gaussian field.}
    \label{fig:overview}
\end{figure*}

\section{Introduction}

Novel view synthesis (NVS) aims to recover the geometry and appearance of a scene from a limited set of images and render views from camera poses that were not observed during training. It has been used in areas such as virtual reality, autonomous driving, and visual effects. Neural radiance fields (NeRFs) ~\cite{1,2,3,4} are able to reconstruct detailed scenes, but their rendering process relies on sampling points along camera rays, which makes inference relatively expensive. 3D Gaussian Splatting (3DGS) ~\cite{5} uses anisotropic Gaussian primitives to represent a scene and performs differentiable rasterization, offering a practical compromise between rendering quality and computational cost.  

Under sparse-view supervision, limited geometric constraints make 3DGS susceptible to overfitting, producing blurred regions, floating artifacts, or distorted structures at unseen viewpoints ~\cite{6}. Existing approaches incorporate depth or geometric priors ~\cite{7,8}, image- or video-diffusion guidance ~\cite{9,10}, and modifications to Gaussian representations ~\cite{11,12}. Dropout-based methods randomly deactivate Gaussians during training ~\cite{13,14,15}, while structured masking strategies further consider spatial anchors, paired consistency, or depth reliability ~\cite{16,17,18}. However, the lineage relationships formed during densification remain less explored in dropout design.  

Dropout changes the Gaussian subsets participating in rendering, reducing dependence on fixed combinations and potentially improving generalization. However, neighboring Gaussians may compensate for dropped primitives, weakening the effect of independent dropout. We define a lineage family as an initialized Gaussian, termed the founder, together with its descendants generated through cloning or splitting. Members of the same family often exhibit spatial clustering and contribute to nearby regions. Independent dropout may therefore leave many related members active, limiting its effect on the family's overall contribution.  

Based on these observations, we propose Stage-wise Lineage-aware Dropout (SLD), which applies shared masks to the descendants of selected families during the structure-regularization stage. Because family-level deactivation may interfere with late-stage detail optimization, SLD subsequently switches to independent Gaussian dropout. We further introduce Lineage-Balanced Densification (LBD), which constructs an expanded candidate pool and redistributes fixed cloning and splitting quotas according to family-level mean gradient scores and existing descendant counts. 

The main contributions of this work are summarized as follows:

\begin{itemize}

\item We investigate the family structure that develops during Gaussian densification in sparse-view Gaussian Splatting. The analysis reveals substantial variation among lineage families in their sizes, generation depths, and spatial arrangements. This variation supports treating lineage families as useful structural units for studying regularization and densification growth.

\item We propose Stage-wise Lineage-aware Dropout (SLD), which applies family dropout to regulate the participation of related Gaussian branches during optimization. We also introduce Lineage-Balanced Densification (LBD), which forms an expanded candidate pool across different families and redistributes the cloning and splitting budgets according to family-level gradient score and the number of existing descendants.

\item We perform quantitative and qualitative comparisons, module-level ablations, and cross-framework integration experiments on several sparse-view benchmarks. The results show competitive performance under the tested settings and provide further evidence for the different contributions of SLD and LBD.

\end{itemize}

\begin{figure}[t]
    \centering
    \includegraphics[width=\columnwidth]{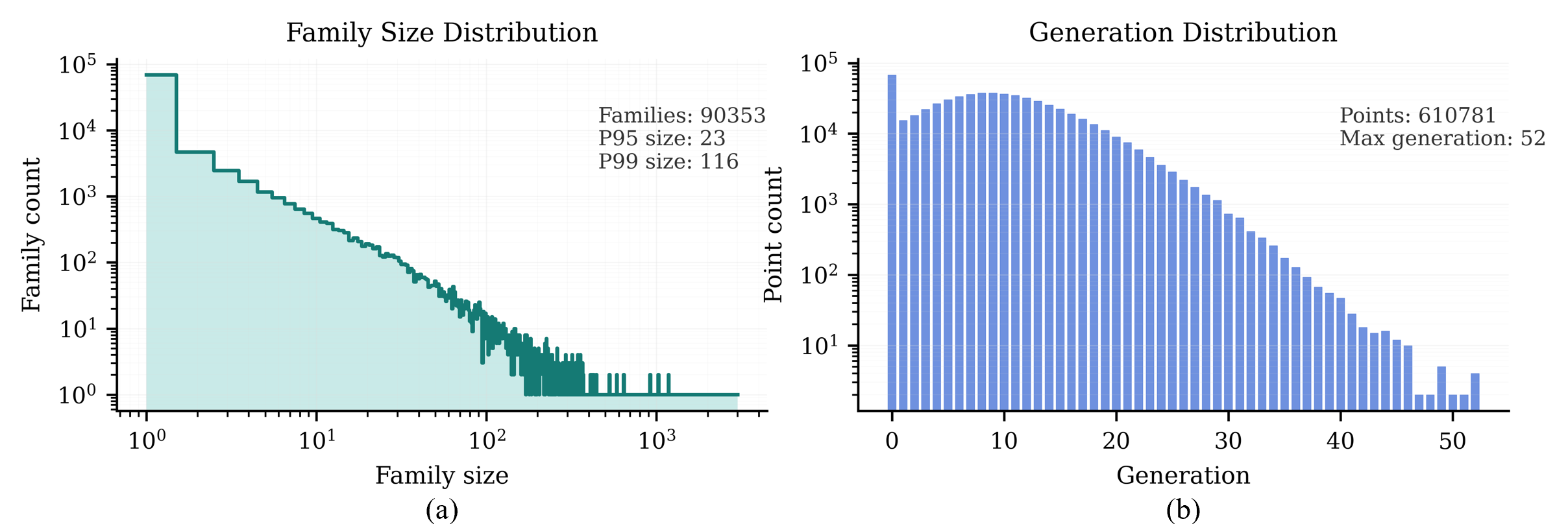}
    \caption{Statistical characterization of densification-induced lineage families: (a) Family-size distribution; (b) Generation distribution.}
    \label{fig:pilot-stats}
\end{figure}

\begin{figure*}[t]
    \centering
    \includegraphics[width=\textwidth]{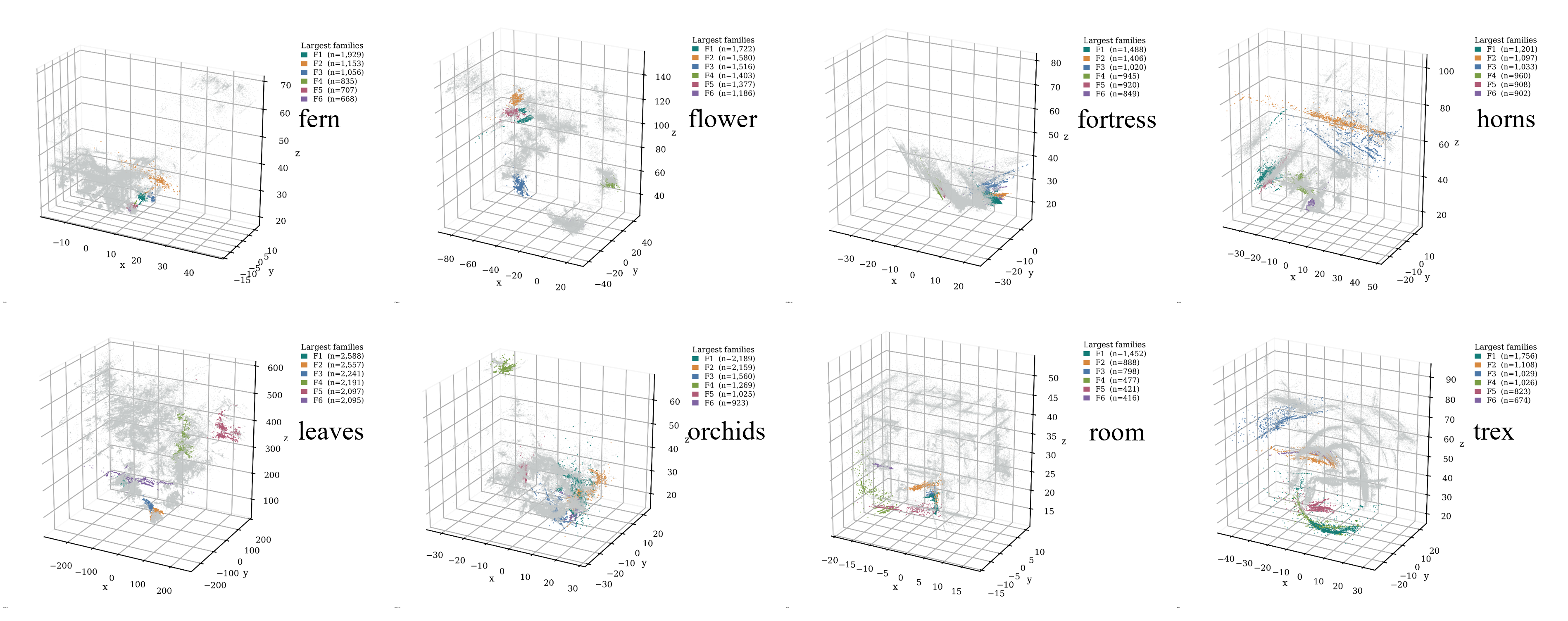}
    \caption{Spatial distribution of lineage families across multiple scenes: The six largest families in each scene are colored, and the remaining Gaussians are shown in gray.}
    \label{fig:pilot-spatial}
\end{figure*}

\begin{figure}[t]
    \centering
    \includegraphics[width=\columnwidth]{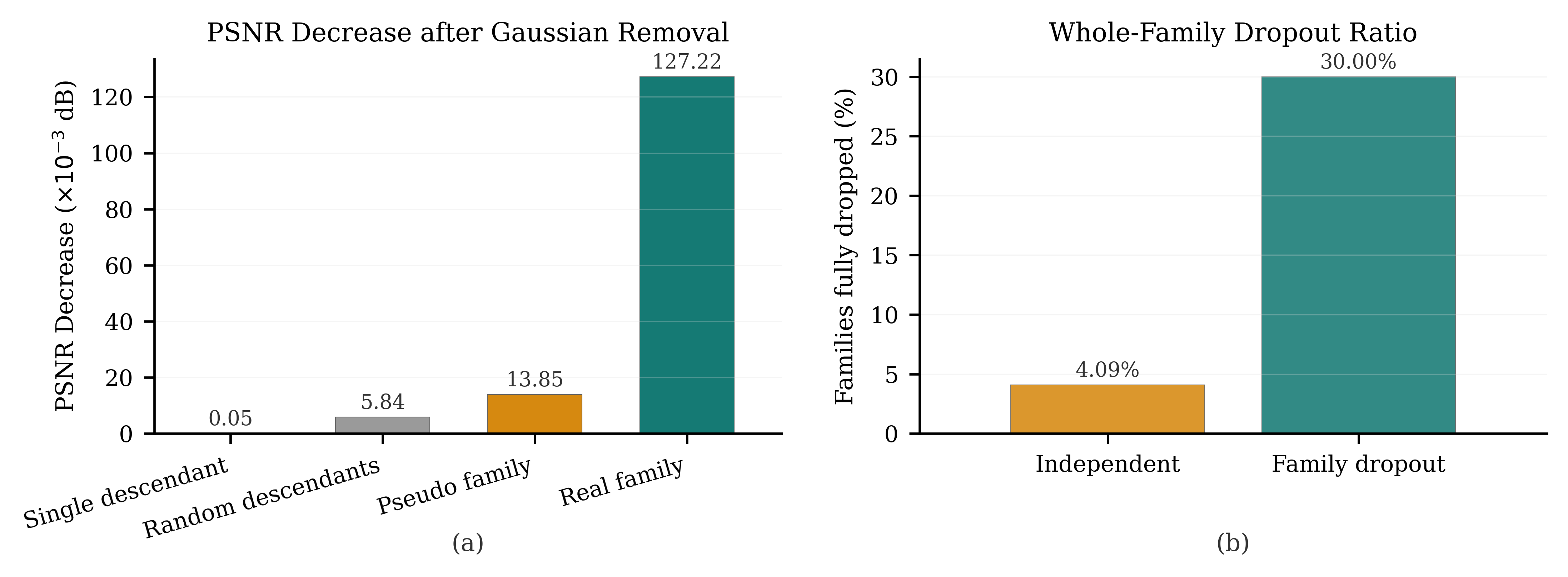}
    \caption{Study of lineage-aware dropout: (a) Probe for all kinds of dropouts. Based on the 10,000 iterations checkpoint of DropGaussian on the LLFF dataset, four removal tests for each LLFF scene are performed: all descendants in the largest lineage family (with count d0), d0 random descendants from top-5 largest families, d0 random Gaussians, and one random descendant from the largest family. The height of bars indicates the PSNR reduction of rendered images before and after one-step of Gaussian removal. (b) Explore the relationship between family dropout and individual Gaussian dropout at the same dropout rate. The height of orange bar indicates the corresponding family dropout rate given the same individual Gaussian dropout rate.}
    \label{fig:pilot-evidence}
\end{figure}

\section{Related Work}

\subsection{Sparse-View 3D Gaussian Splatting}

NeRFs ~\cite{1} and related radiance-field representations ~\cite{2,3,4,19} have advanced novel view synthesis, with sparse-view methods introducing depth priors, semantic guidance, and cross-view regularization ~\cite{20,21,22,23,24,25,26,27}. However, ray-based volume rendering remains computationally demanding. 3DGS ~\cite{5} provides an efficient explicit alternative through differentiable rasterization, but limited observations can still lead to overfitting and geometric artifacts ~\cite{6}.

Sparse-view 3DGS methods incorporate depth or geometric priors ~\cite{7,8} and neighborhood-aware point-cloud unpooling ~\cite{34}. Related developments explore alternative representations and structural constraints ~\cite{35,36,37}, as well as Gaussian refinement and compression ~\cite{11,12}. Beyond static scenes, dynamic Gaussian representations and efficient/compact variants have also been explored ~\cite{28,29,30,31,32,33}. These approaches improve scene representations, while the ancestry relationships created during densification remain less explored as a basis for regularization.

\subsection{Dropout-Based Sparse Regularization}

Dropout reduces co-adaptation in deep networks ~\cite{38,39,40,41} and has been adapted to sparse-view 3DGS. DropGaussian ~\cite{14} and DropoutGS ~\cite{15} randomly deactivate Gaussians during training to alleviate overfitting. More structured methods guide masking through spatial or reliability information. DropAnSH-GS ~\cite{16} combines anchor-based dropout with spherical-harmonic restrictions, PairDropGS ~\cite{17} introduces paired dropout and low-frequency consistency, and DOC-GS ~\cite{18} uses dark-channel and depth information to identify reliable Gaussians. A related form of attribute masking has also been used in reflective appearance modeling ~\cite{42}.

These methods primarily organize masking around individual Gaussians or spatial neighborhoods rather than ancestry relationships. In contrast, our method groups Gaussians by their shared founder and applies family-shared masks to related descendants. It subsequently switches to independent dropout, incorporating both lineage structure and training stage into the regularization design.

\section{Method}
In this work, we propose a lineage-aware framework for sparse-view 3DGS combining Stage-wise Lineage-aware Dropout (SLD) and Lineage-Balanced Densification (LBD). SLD controls which Gaussians participate in rendering, while LBD redistributes fixed cloning and splitting quotas among lineage families. The overall framework is shown in Fig.~\ref{fig:overview}.

\subsection{Preliminaries}

3D Gaussian Splatting (3DGS) represents a 3D scene using a set of anisotropic Gaussian primitives. Each Gaussian is explicitly parameterized by its center position, 3D covariance (representing scale and rotation), opacity, and Spherical Harmonics (SH) coefficients for view-dependent appearance. Given a specific camera viewpoint, 3DGS utilizes differentiable tile-based rasterization to project these 3D Gaussians onto the 2D image plane. The final color $C$ of an arbitrary pixel is obtained by accumulating the ordered Gaussians via alpha-blending:

\begin{equation}
C = \sum_{i=1}^{N} T_i\alpha_i C_i.
\label{eq:alpha-blending}
\end{equation}

where

\begin{equation}
T_i = \prod_{j=1}^{i-1}(1-\alpha_j).
\label{eq:transmittance}
\end{equation}
Here, $C_i$ and $\alpha_i$ denote the color and opacity of the $i$-th Gaussian, and $T_i$ is the cumulative transmittance of the Gaussians in front of it.

\subsection{Pilot Study}

This pilot study analyzes lineage-family sizes, generation depths, and spatial distributions, and compares Gaussian removal schemes to inform dropout design. Experiments follow the DropGaussian~\cite{14} protocol with three input views per LLFF scene, and lineage statistics are collected from a single training run.

Fig.~\ref{fig:pilot-stats}(a) shows that family sizes vary widely. The statistics on the number of family members reveal that the sizes of most families are small, whereas a small number of families possess many members: 95\% of families contain no more than 23 members, and 99\% of families contain no more than 116 members. Fig.~\ref{fig:pilot-stats}(b) shows the counts of Gaussian primitives at different generations through cloning and splitting operations.

Besides, the spatial distribution of families in eight LLFF scenes as shown in Fig.~\ref{fig:pilot-spatial}. Across eight scenes, the top six largest families are colored with identical colors respectively. In some scenes, their members are gathered in small areas. In others, they spread over longer or wider regions. Even so, the members of most families are not randomly scattered across the entire scene. The spatial distribution of lineage families exhibits a certain degree of spatial clustering.

We also compare different dropout strategies, and their effects on performance are shown in Fig.~\ref{fig:pilot-evidence}(a). Deleting only one descendant brings negligible changes, whereas randomly removing Gaussians leads to more visible degradation. Removing descendants from multiple lineage families causes even stronger degradation, and removing an entire lineage family results in the largest performance drop. This suggests that Gaussians from the same family cooperate to construct the scene representation. Fig.~\ref{fig:pilot-evidence}(b) explores the relationship between family dropout and individual Gaussian dropout at the same dropout rate. Family dropout deactivates a complete family, whereas individual dropout often removes only part of a family, since the chance of dropping all descendants at the same time is low, around 4.09\% in our experiment.

The spatial distribution and removal analysis of lineage family members support using lineage families as the basic unit for structured dropout.

\subsection{Stage-wise Lineage-aware Dropout}
Based on this analysis, we propose Stage-wise Lineage-aware Dropout (SLD), which jointly deactivates descendants of selected families during the structure-regularization stage. Gradient-driven densification responds to local geometry, appearance, and visibility, so family members may cluster along local surfaces. Family-shared dropout aims to reduce reliance on fixed combinations of these spatially related Gaussians.

\noindent\textbf{Lineage Partition.} During densification, each Gaussian stores a founder ID and a generation index. Initial Gaussians are assigned generation zero. For a new Gaussian generated by cloning or splitting, its founder ID is inherited from its parent, and its generation index is set to one greater than that of the parent. Gaussians with the same founder ID form one lineage family.

\noindent\textbf{Stage-wise Scheduling and Masking.} To balance structural regularization and detail recovery, we divide training into three stages: warm-up, structure regularization, and detail refinement. Let $t$ denote the current iteration, $t_w$ the end of the warm-up stage, and $t_s$ the transition point to detail refinement. The dynamic dropout rate $p_t$ is formally defined as:

\begin{equation}
p_t = \begin{cases}
0, & t \le t_w,\\[3pt]
p_{\max}\dfrac{t-t_w}{t_s-t_w}, & t_w < t < t_s,\\[6pt]
p_3, & t \ge t_s,
\end{cases}
\label{eq:dropout-schedule}
\end{equation}
where $p_{\max}$ is the maximum dropout rate during the structure regularization phase, and $p_t$ is the fixed rate during the detail refinement phase. No dropout is applied during the initial phase of training ($t \le t_w$). During the middle stage, lineage families are dropped with probability $p_t$. For each dropped family, descendants with generation indices greater than zero are added to the dropout set $D$. Adopting this founder-preserving policy, generation-0 Gaussians are excluded from dropout. Unlike vanilla 3DGS, the founders are also retained when pruning or splitting during the first two stages. After $t_s$, whole family dropout is replaced by independent Gaussian dropout. Since then, all Gaussians, including founders, are subject to drop with probability $p_3$.

A binary mask \(m_i\) is assigned to each Gaussian \(G_i\) according to the dropout set \(D\).

\begin{equation}
m_i = \begin{cases}
0, & G_i\in D,\\
1, & G_i\notin D.
\end{cases}
\label{eq:dropout-mask}
\end{equation}
During the forward rasterization pass, the original opacity $\alpha_i$ of each Gaussian $G_i$ is modulated by this mask and compensation factor to obtain the effective opacity $\hat{\alpha}_i$:

\begin{equation}
\hat{\alpha}_i = \begin{cases}
\alpha_i, & G_i \text{ is not eligible for dropout},\\[3pt]
\dfrac{m_i\alpha_i}{1-p_t}, & G_i \text{ is eligible for dropout}.
\end{cases}
\label{eq:effective-opacity}
\end{equation}
The compensation factor $\frac{1}{1 - p_t}$ is applied to the opacities of the remaining Gaussians that are dropout-eligible, whereas the opacities of the dropped Gaussians are assigned to zero. During the second stage, protected Gaussians, i.e. generation-0 founders, are excluded from this scaling of opacity and retain their original opacities. During the detail refinement stage with independent dropout, the founders are eligible for dropout and their opacities are also rescaled.

\begin{figure*}[t]
    \centering
    \includegraphics[width=\textwidth]{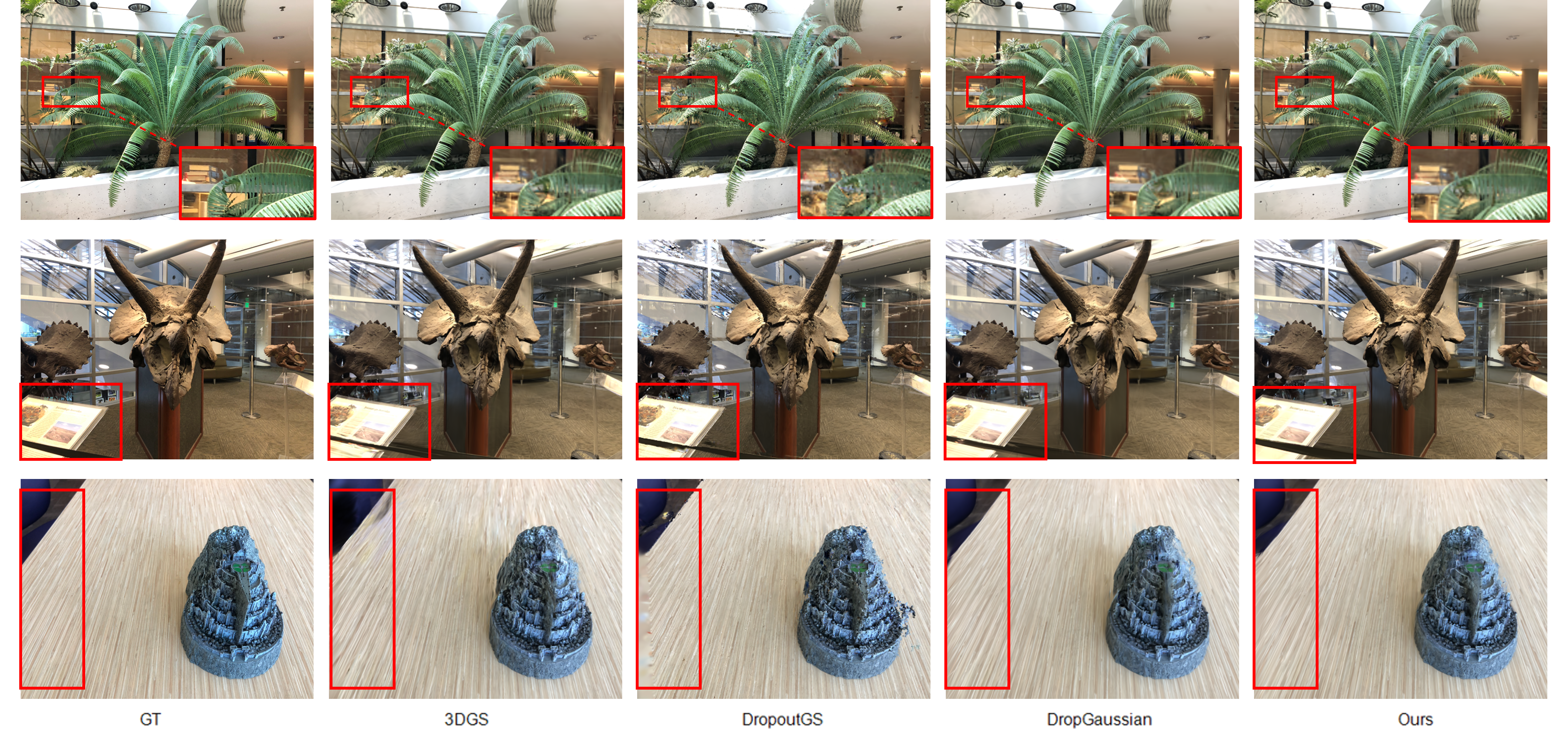}
    \caption{Qualitative comparison on the LLFF dataset. From left to right, the columns show the ground truth, 3DGS, DropoutGS, DropGaussian, and the proposed method. The boxed regions focus on fine structures, object boundaries, and local textures.}
    \label{fig:qual-llff}
\end{figure*}

\begin{table*}[t]
\centering
\caption{Quantitative comparison on the LLFF dataset. The best results are highlighted in bold, the second-best results are underlined, and the third-best results are marked with dotted underlines.}
\label{tab:llff}
\footnotesize
\setlength{\tabcolsep}{3.5pt}
\begin{tabular*}{\textwidth}{@{\extracolsep{\fill}}llccccccccc}
\toprule
\multirow{2}{*}{Category} & \multirow{2}{*}{Method} & \multicolumn{3}{c}{3-view} & \multicolumn{3}{c}{6-view} & \multicolumn{3}{c}{9-view} \\
\cmidrule(lr){3-5} \cmidrule(lr){6-8} \cmidrule(lr){9-11}
 & & PSNR$\uparrow$ & SSIM$\uparrow$ & LPIPS$\downarrow$ & PSNR$\uparrow$ & SSIM$\uparrow$ & LPIPS$\downarrow$ & PSNR$\uparrow$ & SSIM$\uparrow$ & LPIPS$\downarrow$ \\
\midrule
\multirow{5}{*}{NeRF-based} & Mip-NeRF~\cite{4} & 16.11 & 0.401 & 0.460 & 22.91 & 0.756 & 0.213 & 24.88 & 0.826 & 0.170 \\
 & DietNeRF~\cite{20} & 14.94 & 0.370 & 0.496 & 21.75 & 0.717 & 0.248 & 24.28 & 0.801 & 0.183 \\
 & RegNeRF~\cite{21} & 19.08 & 0.587 & 0.336 & 23.10 & 0.760 & 0.206 & 24.86 & 0.820 & 0.161 \\
 & FreeNeRF~\cite{26} & 19.63 & 0.612 & 0.308 & 23.73 & 0.779 & 0.195 & 25.13 & 0.827 & 0.160 \\
 & SparseNeRF~\cite{27} & 19.86 & 0.624 & 0.328 & -- & -- & -- & -- & -- & -- \\
\midrule
\multirow{7}{*}{3DGS-based} & 3DGS~\cite{5} & 19.17 & 0.646 & 0.268 & 23.74 & 0.807 & 0.162 & 25.44 & 0.860 & 0.096 \\
 & DNGaussian~\cite{43} & 19.12 & 0.591 & 0.294 & 22.18 & 0.755 & 0.198 & 23.17 & 0.788 & 0.180 \\
 & FSGS~\cite{34} & \dotuline{20.43} & 0.682 & 0.248 & 24.09 & 0.823 & 0.145 & 25.31 & 0.860 & 0.122 \\
 & CoR-GS~\cite{44} & 20.36 & \dotuline{0.710} & \dotuline{0.202} & 24.34 & \dotuline{0.831} & \dotuline{0.122} & \dotuline{25.94} & \underline{0.872} & \underline{0.088} \\
 & DropoutGS~\cite{15} & 19.39 & 0.632 & 0.279 & 24.02 & 0.816 & 0.144 & 25.13 & 0.869 & 0.099 \\
 & DropGaussian~\cite{14} & 20.33 & 0.709 & \underline{0.201} & \dotuline{24.58} & 0.830 & 0.125 & 25.85 & 0.864 & \dotuline{0.093} \\
 & DropAnSH-GS~\cite{16} & \underline{20.68} & \textbf{0.724} & \textbf{0.194} & \textbf{24.76} & \underline{0.837} & \underline{0.116} & \textbf{26.24} & \textbf{0.875} & \underline{0.088} \\
\midrule
 & \textbf{Ours} & \textbf{20.70} & \underline{0.717} & \textbf{0.194} & \underline{24.72} & \textbf{0.838} & \textbf{0.111} & \underline{26.17} & \textbf{0.875} & \textbf{0.086} \\
\bottomrule
\end{tabular*}
\end{table*}

\begin{figure*}[t]
    \centering
    \includegraphics[width=\textwidth]{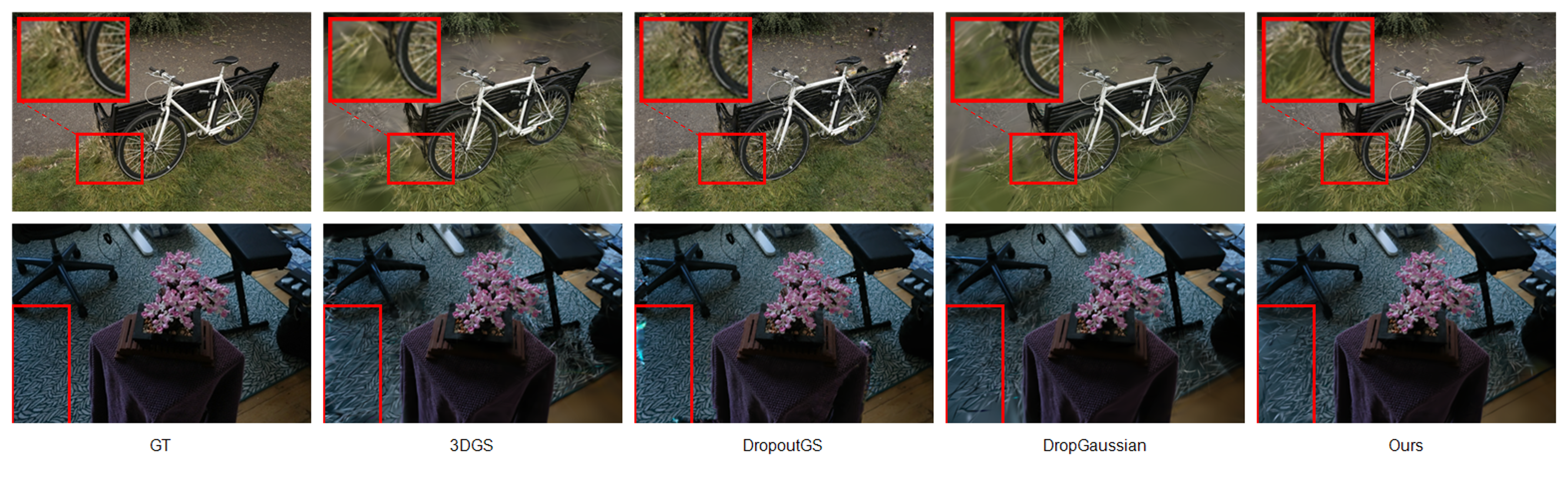}
    \caption{Qualitative comparison on the MipNeRF360 dataset. From left to right, the columns show the ground truth, 3DGS, DropoutGS, DropGaussian, and the proposed method. The boxed regions highlight thin structures, cluttered background and patterned textures.}
    \label{fig:qual-mipnerf360}
\end{figure*}

\begin{figure*}[t]
    \centering
    \includegraphics[width=\textwidth]{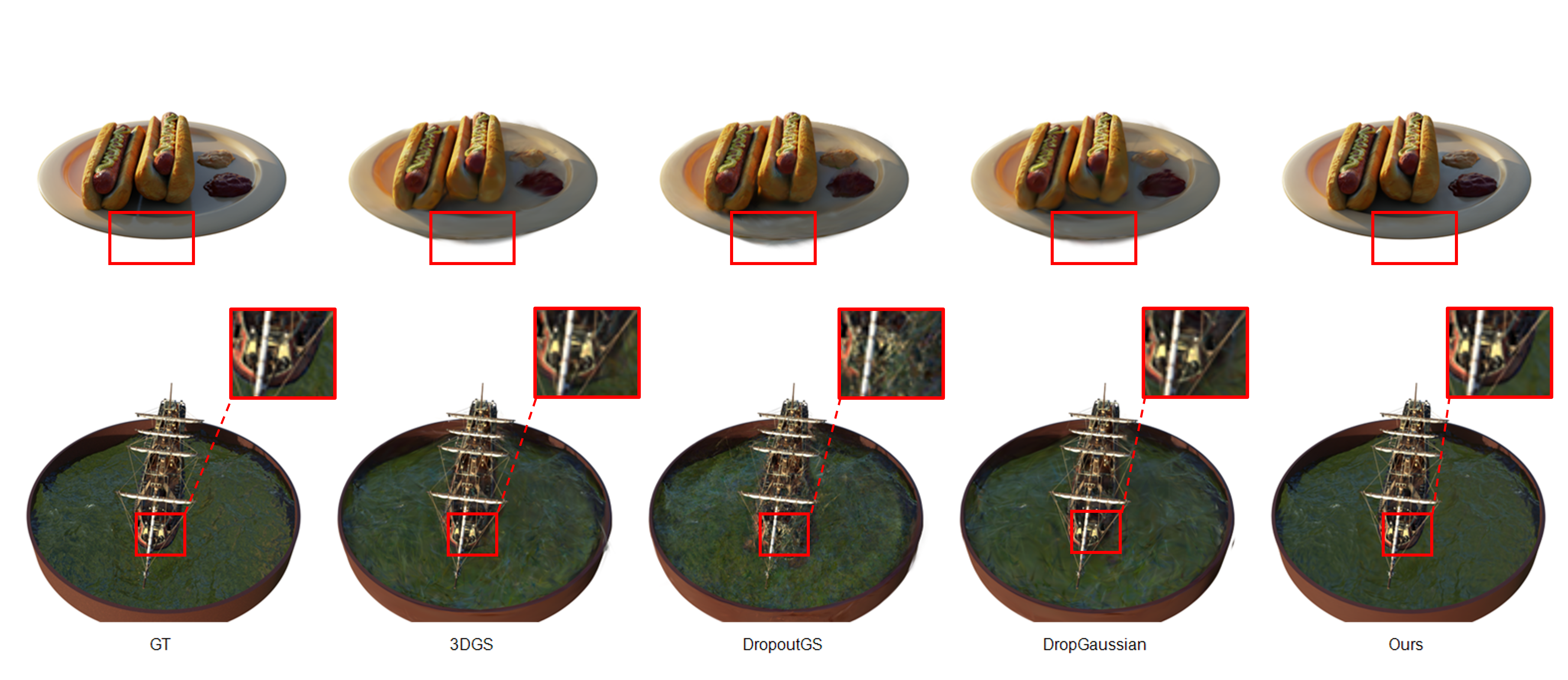}
    \caption{Qualitative comparison on the Blender dataset. From left to right, the columns show the ground truth, 3DGS, DropoutGS, DropGaussian, and the proposed method. The boxed regions focus on object contours and appearance details.}
    \label{fig:qual-blender}
\end{figure*}

\subsection{Lineage-Balanced Densification}
We introduce Lineage-Balanced Densification (LBD) to complement SLD by redistributing densification opportunities according to family-level gradient scores and descendant counts. LBD regulates growth concentration without enforcing equal family sizes, while preserving the cloning and splitting quotas determined by the original 3DGS criteria.

\noindent\textbf{Fixed Budgets and Expanded Candidate Pool.} At each densification step, LBD computes the screen-space positional gradients and applies the original 3DGS gradient threshold $\theta_g$ together with the operation-specific criteria to construct separate candidate sets for cloning and splitting, denoted by $\mathcal{P}_t^{\text{clone}}$ and $\mathcal{P}_t^{\text{split}}$, respectively. Rather than densifying these candidates immediately, LBD uses the size of each set as the fixed budget for its corresponding operation, $K_o = |\mathcal{P}_t^o|$, where $o \in \{\text{clone}, \text{split}\}$. These budgets are then redistributed across lineage families using the corresponding expanded candidate pools. To provide candidates from more lineage families for reallocation, we construct an expanded cross-family candidate pool $\mathcal{P}^o$ using a lower gradient threshold $\theta_{\text{pool}} = \gamma\theta_g$, where $\gamma$ defaults to 0.5. Since the expanded pool uses a lower gradient threshold, it contains all candidates selected under the original threshold. 

\noindent\textbf{Family Growth Score.} Let \(m_{k}^{o}\) denote the number of candidates of lineage family \(k\) in the expanded pool for operation \(o\), and let \(u_{k}^{o}\) denotes the mean gradient score of the candidates, which is defined as

\begin{equation}
u_k^o = \frac{1}{m_k^o} \sum_{i \in \mathcal{P}^o \cap \mathcal{F}_k} s_i,
\end{equation}
where \(s_i\) denotes the gradient value of candidate \(i\). A larger \(u_k^o\) indicates a stronger densification signal.

To balance densification demand and family size, we define the lineage-balanced growth score of family $k$ as

\begin{equation}
w_k^o(t) = \frac{u_k^o(t)}{(d_k(t) + \tau)^{\beta_t}},
\end{equation}
where \(d_k(t)\) denotes the number of descendants in family \(k\) at iteration \(t\), \(\tau\) is a constant 1, and \(\beta_t\) controls the family-size penalty.

In the fixed budget $K_o$, LBD assigns a candidate quota to each family according to:

\begin{equation}
q_k^o = \min\left(m_k^o, \lambda_o w_k^o(t)\right), \quad \sum_k q_k^o = K_o,
\end{equation}
where \(\lambda_o\) scales the family growth scores to obtain provisional quotas, while the minimum operation limits each quota by its candidate capacity \(m_k^o\). We select the candidates with the top \(q_k^o\) highest gradient scores from each family for cloning/splitting operations. The penalty term \(\beta_t\) is gradually increased to \(\beta_{\max}\) during training and remains fixed thereafter.

LBD determines which candidate Gaussians are selected for cloning or splitting during densification, whereas SLD regulates which Gaussians participate in rendering. For more implementation details, see the supplementary materials.

\section{Experiments}

\begin{table}[t]
\centering
\caption{Quantitative comparison on the MipNeRF360 dataset. The best results are highlighted in bold, the second-best results are underlined, and the third-best results are marked with dotted underlines.}
\label{tab:mipnerf360}
\footnotesize
\setlength{\tabcolsep}{3.5pt}
\resizebox{\columnwidth}{!}{%
\begin{tabular}{lcccccc}
\toprule
\multirow{2}{*}{Method} & \multicolumn{3}{c}{12-view} & \multicolumn{3}{c}{24-view} \\
\cmidrule(lr){2-4} \cmidrule(lr){5-7}
 & PSNR$\uparrow$ & SSIM$\uparrow$ & LPIPS$\downarrow$ & PSNR$\uparrow$ & SSIM$\uparrow$ & LPIPS$\downarrow$ \\
\midrule
3DGS~\cite{5} & 18.52 & 0.523 & 0.415 & 22.80 & 0.708 & 0.276 \\
FSGS~\cite{34} & 18.80 & 0.531 & 0.418 & \dotuline{23.70} & \dotuline{0.745} & \dotuline{0.230} \\
CoR-GS~\cite{44} & 19.52 & 0.558 & 0.418 & 23.39 & 0.727 & 0.271 \\
DropGaussian~\cite{14} & \dotuline{19.66} & \dotuline{0.569} & \dotuline{0.374} & \underline{23.71} & \underline{0.752} & \underline{0.227} \\
DropAnSH-GS~\cite{16} & \textbf{19.93} & \underline{0.576} & \underline{0.362} & -- & -- & -- \\
\midrule
\textbf{Ours} & \underline{19.83} & \textbf{0.578} & \textbf{0.360} & \textbf{24.15} & \textbf{0.768} & \textbf{0.211} \\
\bottomrule
\end{tabular}%
}
\end{table}

\begin{table}[t]
\centering
\caption{Quantitative comparison on the Blender dataset. The best results are highlighted in bold, the second-best results are underlined, and the third-best results are marked with dotted underlines.}
\label{tab:blender}
\footnotesize
\begin{tabular*}{\columnwidth}{@{\extracolsep{\fill}}lccc}
\toprule
Method & PSNR$\uparrow$ & SSIM$\uparrow$ & LPIPS$\downarrow$ \\
\midrule
3DGS~\cite{5} & 21.56 & 0.847 & 0.130 \\
DNGaussian~\cite{43} & 24.31 & 0.886 & \underline{0.088} \\
FSGS~\cite{34} & 24.64 & \textbf{0.895} & 0.095 \\
CoR-GS~\cite{44} & 24.32 & \underline{0.892} & \textbf{0.084} \\
DropGaussian~\cite{14} & \dotuline{25.19} & 0.890 & 0.090 \\
DropAnSH-GS~\cite{16} & \textbf{25.50} & \dotuline{0.891} & \underline{0.088} \\
\midrule
\textbf{Ours} & \underline{25.20} & \textbf{0.895} & \dotuline{0.089} \\
\bottomrule
\end{tabular*}
\end{table}

\subsection{Experimental Setup}

\noindent\textbf{Datasets.} We evaluate the proposed framework on three datasets, LLFF, MipNeRF360, and Blender, which contain different scene types and view settings. 

\noindent\textbf{Baselines.} We compare the proposed method with both NeRF-based and 3DGS-based sparse-view reconstruction methods. The NeRF-based baselines include Mip-NeRF~\cite{4}, DietNeRF~\cite{20}, RegNeRF~\cite{21}, FreeNeRF~\cite{26}, and SparseNeRF~\cite{27}. The 3DGS-based baselines include standard 3DGS~\cite{5}, DNGaussian~\cite{43}, FSGS~\cite{34}, CoR-GS~\cite{44}, DropoutGS~\cite{15}, DropGaussian~\cite{14}, and DropAnSH-GS~\cite{16}.

\noindent\textbf{Implementation Details.} Our experiments are based on the official DropGaussian~\cite{14} framework and its training and evaluation settings. Baselines with released codes follow the authors' implementations, while results for other methods are taken from their original papers or public reports. For SLD, whole family dropout starts after warm-up (\(t_w = 3000\)) and converts to independent Gaussian dropout after \(t_s = 7000\). We set \(p_{\max} = 0.3\) and \(p_3 = 0.2\). For LBD, expanded candidate pool coefficient \(\gamma = 0.5\), and \(\beta_t\) increases from 0 to \(\beta_{\max} = 0.5\) between \(t_w\) and \(t_s\). Cloning and splitting operations are processed separately, and each operation keeps its original budget \(K_o\). Our methods and reproduced baselines are averaged over three runs. Results from prior publications follow the original authors' protocols.

\noindent\textbf{Evaluation Metrics.} We use PSNR, SSIM, and LPIPS to evaluate novel-view reconstruction. Higher PSNR and SSIM values and a lower LPIPS value indicate better reconstruction results.

More experimental analyses are provided in the supplementary materials.

\subsection{Quantitative Comparison}

Tables~\ref{tab:llff}, \ref{tab:mipnerf360}, and \ref{tab:blender} present the quantitative results of the proposed framework on the three datasets. Table~\ref{tab:llff} first reports the results on LLFF under sparse-view settings. In the 3-view case with the weakest constraints, our method achieves 20.70~dB in PSNR, which is 0.37~dB higher than DropGaussian~\cite{14}, while its perceptual quality remains competitive. The results across different input-view settings show that our lineage-aware modules achieve competitive PSNR and SSIM metrics (best or second-best), while obtaining the best LPIPS scores, which demonstrates their effectiveness for sparse-view reconstruction.

\begin{table}[t]
\centering
\caption{Ablation of SLD and LBD under 3-view inputs on the LLFF dataset. The best results are highlighted in bold.}
\label{tab:ablation-sld-lbd}
\footnotesize
\begin{tabular*}{\columnwidth}{@{\extracolsep{\fill}}ccccc}
\toprule
\multicolumn{2}{c}{Settings} & \multirow{2}{*}{PSNR$\uparrow$} & \multirow{2}{*}{SSIM$\uparrow$} & \multirow{2}{*}{LPIPS$\downarrow$} \\
\cmidrule(lr){1-2}
SLD & LBD & & & \\
\midrule
$\times$ & $\times$ & 19.17 & 0.646 & 0.268 \\
$\times$ & $\checkmark$ & 19.84 & 0.681 & 0.213 \\
$\checkmark$ & $\times$ & 20.50 & 0.715 & 0.197 \\
$\checkmark$ & $\checkmark$ & \textbf{20.70} & \textbf{0.717} & \textbf{0.194} \\
\bottomrule
\end{tabular*}
\end{table}

\begin{table}[t]
\centering
\caption{Ablation of dropout granularity and founder protection policy under 3-view inputs on the LLFF dataset. The best results are highlighted in bold.}
\label{tab:ablation-granularity}
\footnotesize
\setlength{\tabcolsep}{3pt}
\begin{tabular*}{\columnwidth}{@{\extracolsep{\fill}} >{\centering\arraybackslash}p{0.25\columnwidth} >{\centering\arraybackslash}p{0.25\columnwidth} ccc}
\toprule
\begin{tabular}[c]{@{}c@{}}Dropout granularity\\ in stage 2\end{tabular} & 
\begin{tabular}[c]{@{}c@{}}Founder policy\\ in stage 1 \& 2\end{tabular} & 
PSNR\(\uparrow\) & SSIM\(\uparrow\) & LPIPS\(\downarrow\) \\
\midrule
Individual & droppable & 20.38 & 0.703 & 0.205 \\
Individual & preserved & 20.22 & 0.708 & 0.205 \\
Family     & droppable & 20.30 & 0.704 & 0.204 \\
Family     & preserved & \textbf{20.50} & \textbf{0.715} & \textbf{0.197} \\
\bottomrule
\end{tabular*}
\end{table}

\begin{table}[t]
\centering
\caption{Ablation on \(t_s\) between Stage 2 and Stage 3 under 3-view inputs on the LLFF dataset. The best results are highlighted in bold.}
\label{tab:sensitivity_stage_transition}
\small
\begin{tabular*}{\columnwidth}{@{\extracolsep{\fill}}cccc}
\toprule
\(t_s\) & PSNR$\uparrow$ & SSIM$\uparrow$ & LPIPS$\downarrow$ \\
\midrule
5000 & $20.4284 \pm 0.1145$ & $0.7062 \pm 0.0029$ & $0.2029 \pm 0.0018$ \\
7000 & $20.4329 \pm 0.0519$ & \textbf{$0.7073 \pm 0.0011$} & \textbf{$0.2018 \pm 0.0006$} \\
9000 & \textbf{$20.4509 \pm 0.0123$} & $0.7070 \pm 0.0004$ & $0.2032 \pm 0.0005$ \\
\bottomrule
\end{tabular*}
\end{table}

\begin{table}[t]
\centering
\caption{Cross-framework validation under 3-view inputs on the LLFF dataset. The proposed SLD and LBD modules are added to four sparse-view Gaussian Splatting frameworks. The best results are highlighted in bold.}
\label{tab:cross-framework}
\footnotesize
\begin{tabular*}{\columnwidth}{@{\extracolsep{\fill}}lccc}
\toprule
Methods & PSNR$\uparrow$ & SSIM$\uparrow$ & LPIPS$\downarrow$ \\
\midrule
FSGS~\cite{34} & 20.43 & 0.682 & 0.248 \\
FSGS+Ours & \textbf{20.85} & \textbf{0.719} & \textbf{0.195} \\
CoR-GS~\cite{44} & 20.36 & 0.710 & 0.202 \\
CoR-GS+Ours & \textbf{20.72} & \textbf{0.724} & \textbf{0.194} \\
DNGaussian~\cite{43} & 19.12 & 0.591 & \textbf{0.294} \\
DNGaussian+Ours & \textbf{19.31} & \textbf{0.613} & 0.287 \\
Scaffold-GS~\cite{37} & 18.62 & 0.643 & 0.258 \\
Scaffold-GS+Ours & \textbf{19.28} & \textbf{0.673} & \textbf{0.220} \\
\bottomrule
\end{tabular*}
\end{table}

The results on MipNeRF360 and Blender are shown in Tables~\ref{tab:mipnerf360} and~\ref{tab:blender}. On the MipNeRF360 dataset, our method achieves the highest SSIM and the lowest LPIPS, while ranking second to DropAnSH-GS~\cite{16} in PSNR under the 12-view setting; with 24 input views, it achieves the best results across all three metrics, demonstrating its effectiveness on large real-world scenes with unbounded backgrounds. On the Blender dataset, our method ties with FSGS~\cite{34} for the best SSIM, while it is second only to DropAnSH-GS~\cite{16} in PSNR and ranks third in LPIPS.

\subsection{Qualitative Comparison}
Figures~\ref{fig:qual-llff}, \ref{fig:qual-mipnerf360}, and \ref{fig:qual-blender} show qualitative comparisons on the LLFF, MipNeRF360, and Blender datasets. The compared methods are 3DGS, DropoutGS, DropGaussian, and our method. The boxed areas are used to compare local structures, object boundaries, and texture details more obviously.

In the following examples, our method typically yields sharper object boundaries, more precise local appearance, and finer geometric structures. For LLFF, the contrast is mainly visible in fine geometric details, object boundaries, and textures. For MipNeRF360, the differences are mainly found in thin structures and accurate texture patterns. For Blender, the boxed areas show differences in object contours. Some baseline methods produce local blur, broken structures, or distorted textures, while our results are relatively more stable in these regions.

\subsection{Ablation Studies}
The individual and joint effects of SLD and LBD are reported in Table~\ref{tab:ablation-sld-lbd}. The results show that both SLD and LBD improve the reconstruction quality, with SLD yielding a notable gain when used alone. The combination of SLD and LBD yields further improvements, indicating that the two modules provide complementary benefits during training. SLD controls which Gaussians participate in rendering, while LBD distributes densification opportunities among different lineage families.

The effects of individual dropout and family-shared dropout under different founder policies are shown in Table~\ref{tab:ablation-granularity}. When enabled, protection retains founders during pruning and splitting in Stages 1 and 2 and excludes them from dropout in Stage 2; it is disabled in Stage 3. With individual dropout, founder preservation improves SSIM but reduces PSNR, leaving LPIPS unchanged. With family-shared dropout, it improves all three metrics and yields the best overall performance, supporting the default SLD configuration.

Table~\ref{tab:sensitivity_stage_transition} shows only small performance differences across \(t_s \in \{5000, 7000, 9000\}\), with all other settings fixed. Although \(t_s=9000\) achieves the highest PSNR, we adopt \(t_s=7000\), which yields the best SSIM and LPIPS.

\subsection{Cross-Framework Validation }
We also test the adaptability of the proposed method by adding the lineage-aware modules to FSGS~\cite{34}, CoR-GS~\cite{44}, DNGaussian~\cite{43} and Scaffold-GS~\cite{37}. After the integration, PSNR and SSIM increase, and LPIPS decreases in all four frameworks, as shown in Table~\ref{tab:cross-framework}. These results indicate that SLD and LBD can be introduced into different sparse-view Gaussian Splatting pipelines, while the extent of improvement depends on the specific framework and evaluation metric.

\section{Conclusion}
We present a lineage-aware framework for sparse-view 3D Gaussian Splatting that combines Stage-wise Lineage-aware Dropout (SLD) with Lineage-Balanced Densification (LBD). SLD applies family-shared dropout before switching to independent dropout, while LBD redistributes densification opportunities among families without changing the original cloning and splitting quotas. Lineage statistics and ablation studies support the use of family information for regularization, with SLD providing the larger gain and LBD offering additional improvements. Quantitative and qualitative comparisons show competitive performance, although the gains vary across datasets, metrics, and training frameworks.

\section*{Acknowledgements}
This work was supported in part by the Joint International Research Laboratory of Spatio-temporal Information and Intelligent Location Services (Grant No. C25GAH31).

\section*{Competing interests}
The authors declare that they have no competing interests or financial conflicts to disclose.

\bibliographystyle{preprintrefs}
\bibliography{ref}
\clearpage
\section*{Supplementary Material}

\section{Complete Descendant Deactivation}

Here we provide the formula for computing the complete-descendant deactivation probability in Pilot Study. We group the Gaussians by founder ID; \(d_k\) denotes the number of descendants in family k, excluding the founders. We equally average the independent dropout probabilities across eligible families within each scene, and then equally average across scenes:

\begin{equation}
R = \frac{1}{|\mathcal{S}|}\sum_{s\in\mathcal{S}}
\frac{1}{|\mathcal{F}_s|}\sum_{k\in\mathcal{F}_s}p^{d_k},
\end{equation}
where \(\mathcal{S}\) denotes the set of scenes and \(\mathcal{F}_s\) the set of lineage families in scene \(s\). If family \(F_k\) has \(d_k\) descendants, all of which are deactivated with probability \(p^{d_k}\).

At \(p=0.30\), independent dropout is estimated to deactivate all eligible descendants in 4.09\% of families, based on their final structures. This is a theoretical estimate, not a measured training-time mask ratio. Under family-shared dropout, the corresponding probability is 30.00\%.

\section{LBD Implementation}

Cloning and splitting are processed separately, with \(o\) denoting the operation. At the current densification step, \(\mathcal{P}_t^o\) is the original candidate set and \(\mathcal{P}^o\) is the expanded pool. For family \(k\), define \(\mathcal{C}_k^o\) as its candidates in the expanded pool:

\begin{equation}
\mathcal{C}_k^o=\mathcal{P}^o\cap\mathcal{F}_k,\;\;
m_k^o=|\mathcal{C}_k^o|,\;\; K_o=|\mathcal{P}_t^o|.
\end{equation}
Thus, \(m_k^o\) is the candidate capacity, whereas \(K_o\) is the fixed number of parent Gaussians to select. The growth score \(w_k^o(t)\) follows the main paper. All calculations below are performed independently for each operation and densification step.

\noindent\textbf{Continuous quotas and numerical solution.} In Eq.~(8) of the main text, \(q_k^o\) in the allocation formula denotes the continuous quota \(\tilde{q}_k^o\). The subsequent top-\(q_k^o\) selection refers to the integer quota obtained after rounding. For clarity, we distinguish the continuous quota \(\tilde{q}_k^o\) from the final integer quota \(q_k^o\) here. A common scaling factor \(\lambda_o\) is chosen so that

\begin{equation}
\widetilde{q}_k^o=\min\!\left(m_k^o,\lambda_o w_k^o(t)\right),\;\;
\sum_k\widetilde{q}_k^o=K_o.
\end{equation}
Zero or full-capacity budgets are assigned directly. If the sum of all family growth scores is at most \(\varepsilon\), uniform weights are used. Otherwise, the monotonic capped-quota sum is solved by 32 bisection updates, starting from a zero lower bound and an upper bound defined as:
\begin{equation}
\max_{k} \frac{m_k^o}{\max(w_k^o(t), \varepsilon)},\;\;
\varepsilon = 10^{-12}.
\end{equation}
The final upper endpoint is used in the quota equation.

\noindent\textbf{Integer quota allocation.} Each continuous quota is rounded down and its fractional remainder is retained:

\begin{equation}
a_k^o=\left\lfloor\widetilde{q}_k^o\right\rfloor,\;\;
r_k^o=\widetilde{q}_k^o-a_k^o.
\end{equation}
We assign the remaining slots one at a time to eligible families, starting with the largest fractional remainder. A family is eligible only when \(a_k^o<m_k^o\). If finite-precision effects cause the integer quotas to differ from \(K_o\), we recompute the capped allocation in float64 and adjust the residual up or down without exceeding family capacities. If the quotas already sum to \(K_o\), we leave them unchanged. The final quotas satisfy:

\begin{equation}
0\leq q_k^o\leq m_k^o,\;\; q_k^o\in\mathbb{Z}_{\geq 0},\;\;
\sum_k q_k^o=K_o.
\end{equation}

\noindent\textbf{Candidate selection within each family.} Finally, the \(q_k^o\) candidates with the highest gradient scores \(s_i\) are selected within each \(\mathcal{C}_k^o\):

\begin{equation}
\mathcal{A}_k^o=\operatorname{Top}_{q_k^o}(\mathcal{C}_k^o;s_i),\;\;
\mathcal{A}^o=\bigcup_k\mathcal{A}_k^o,\;\; |\mathcal{A}^o|=K_o.
\end{equation}
Gradient ties within a family retain the existing candidate order, and the selected parents are cloned or split. The quota counts selected parent Gaussians, not the final number of Gaussians.

\section{Family Growth Analysis}

\begin{figure}[t]
  \centering
  \includegraphics[width=0.45\textwidth]{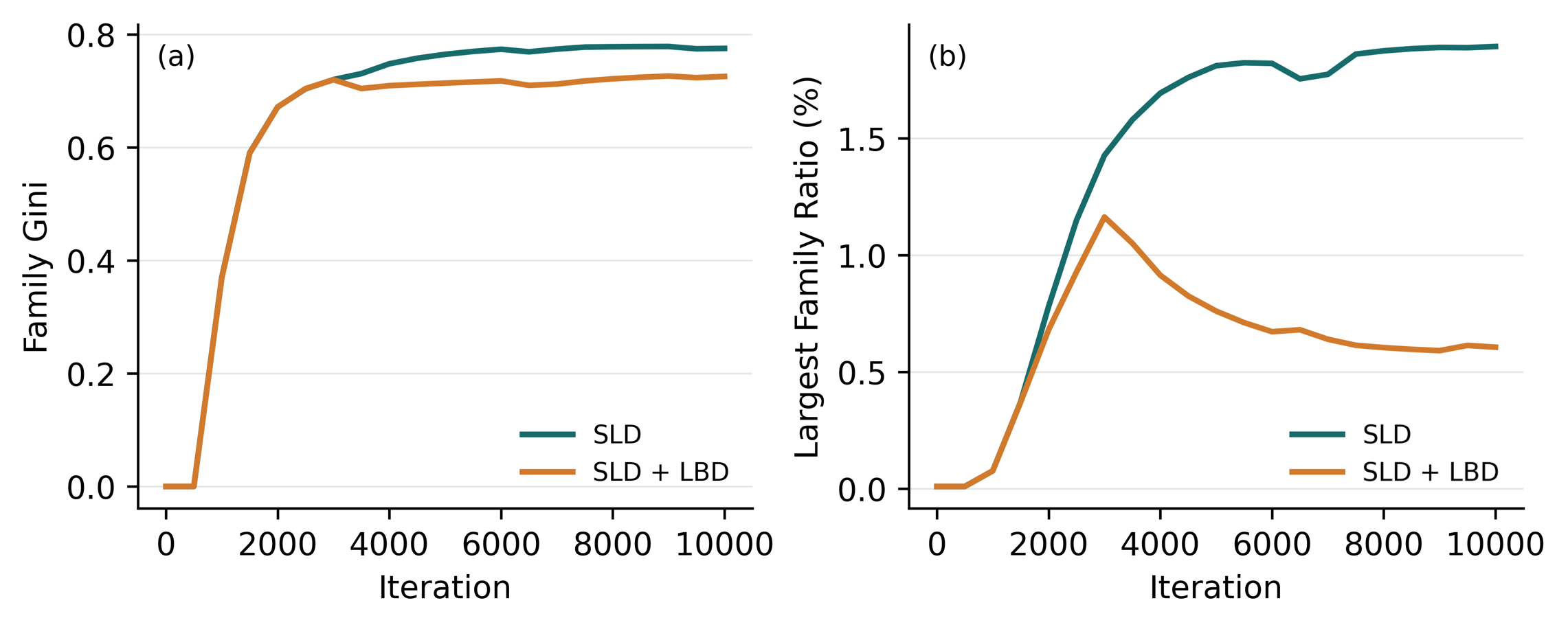}
  \caption{Evolution of lineage-family size concentration during training on LLFF. (a) Family-size Gini coefficient. (b) Percentage of Gaussians belonging to the largest family. SLD and SLD+LBD are compared throughout training.}
  \label{fig:family-evolution}
\end{figure}

\begin{figure}[t]
  \centering
  \includegraphics[width=0.55\linewidth]{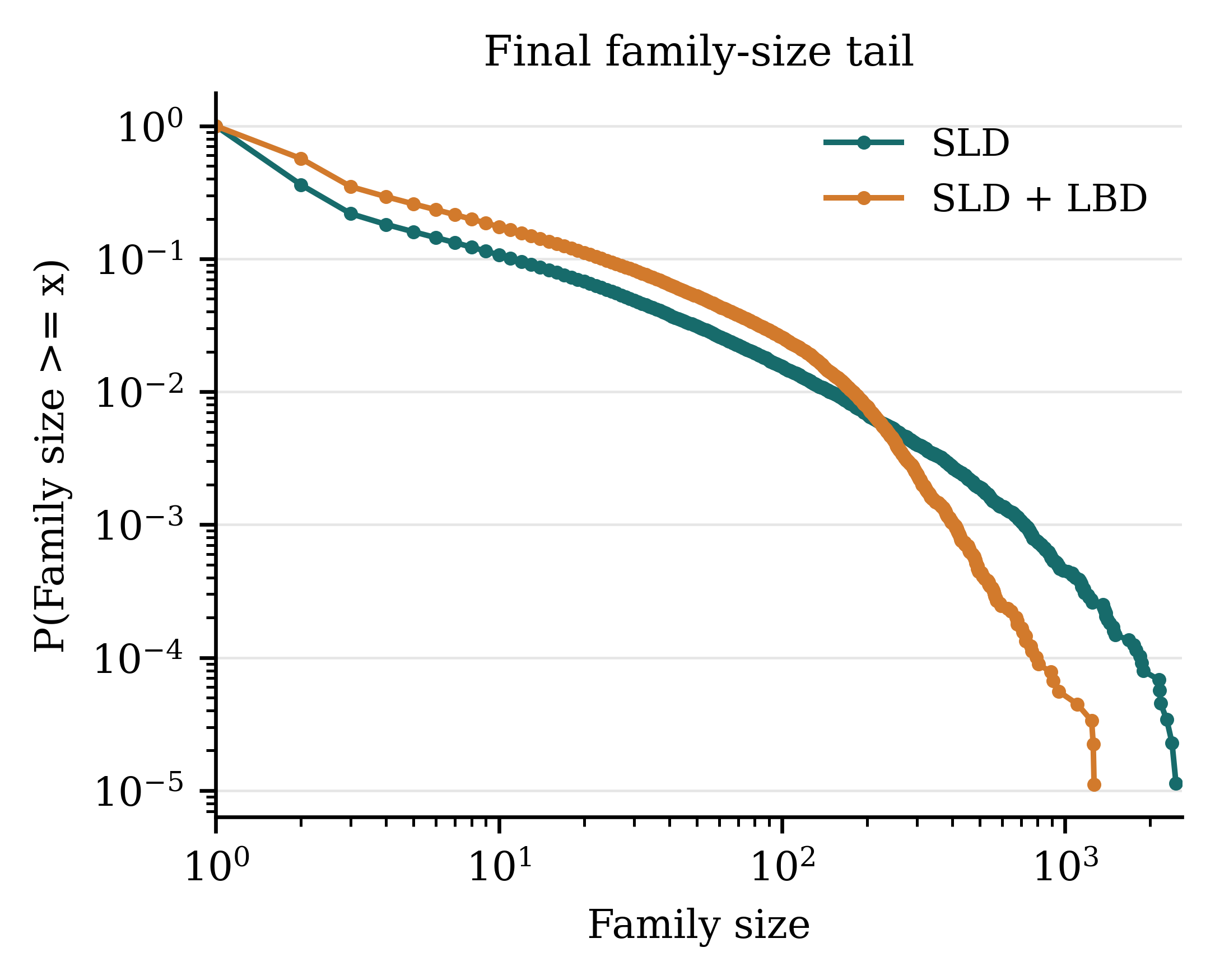} 
  \caption{Final lineage-family size distribution on LLFF. The complementary cumulative distribution is pooled across the eight scenes, and both axes use logarithmic scales. Adding LBD produces a lighter extreme upper tail.}
  \label{fig:family-tail}
\end{figure}

We compare SLD with SLD+LBD on LLFF to examine how LBD changes the distribution of Gaussians across lineage families. Fig.~\ref{fig:family-evolution}(a) reports the family-size Gini coefficient, where a lower value indicates less inequality. Fig.~\ref{fig:family-evolution}(b) reports the percentage of Gaussians belonging to the largest family. Compared with SLD alone, adding LBD produces a lower mean Gini coefficient during later training and reduces the largest family's fraction of the scene's Gaussians.

Fig.~\ref{fig:family-tail} shows the complementary cumulative distribution of final family sizes, pooled across the eight LLFF scenes, with both axes on logarithmic scales. The SLD+LBD curve is higher over some small and intermediate sizes but decreases more rapidly in the extreme upper tail. This suggests that LBD does not uniformly shrink every family; rather, it changes the distribution and reduces the prevalence of extremely large families.

These observations support the role of LBD in moderating concentrated family growth. They do not imply that large families are inherently harmful or that family sizes should be equal, since different regions may require different representation capacities.

\section{Future Directions}
Future work could examine lineage-aware modeling for sparse dynamic (4D) scene reconstruction. As a first step, we can track lineage families across frames and study how their changes relate to scene motion and visibility. This analysis could reveal whether lineage provides structural information over time that is not apparent when frames are considered separately. These findings could then guide the design of spatio-temporal regularization based on lineage relationships and help determine whether it improves reconstruction quality and consistency across viewpoints and time under sparse-view supervision.

\end{document}